\documentclass[11pt,a4paper]{article}
\usepackage[hyperref]{acl2021}
\usepackage{times}
\usepackage{latexsym}
\usepackage{graphicx} 
\usepackage{multirow}
\usepackage{booktabs}
\usepackage{algorithm}
\usepackage{algorithmicx}  
\usepackage{algpseudocode}
\usepackage{url}
\newcommand{\mathbbm}[1]{\text{\usefont{U}{bbm}{m}{n}#1}}

\usepackage{amsmath,amsfonts,bm}

\def\eqref#1{equation~\ref{#1}}

\def\1{\bm{1}}

\def\mH{{\bm{H}}}

\def\mK{{\bm{K}}}

\def\mQ{{\bm{Q}}}

\def\mV{{\bm{V}}}
\def\mW{{\bm{W}}}

\DeclareMathAlphabet{\mathsfit}{\encodingdefault}{\sfdefault}{m}{sl}
\SetMathAlphabet{\mathsfit}{bold}{\encodingdefault}{\sfdefault}{bx}{n}

\def\gA{{\mathcal{A}}}

\newcommand{\R}{\mathbb{R}}

\newcommand{\softmax}{\mathrm{softmax}}

\DeclareMathOperator*{\argmax}{arg\,max}
\DeclareMathOperator*{\argmin}{arg\,min}

\usepackage{microtype}

\aclfinalcopy 
\def\aclpaperid{2316} 

\usepackage{listings}
\usepackage{color}

\definecolor{dkgreen}{rgb}{0,0.6,0}
\definecolor{gray}{rgb}{0.5,0.5,0.5}
\definecolor{mauve}{rgb}{0.58,0,0.82}

\usepackage[normalem]{ulem}

\title{AutoTinyBERT: Automatic Hyper-parameter Optimization \\for Efficient Pre-trained Language Models}

\author{Yichun Yin\textsuperscript{1}, Cheng Chen\textsuperscript{2*}, Lifeng Shang\textsuperscript{1}, Xin Jiang\textsuperscript{1},  Xiao Chen\textsuperscript{1}, Qun Liu\textsuperscript{1}\\ 
 \textsuperscript{1}Huawei Noah’s Ark Lab \\ 
 \textsuperscript{2}Department of Computer Science and Technology, Tsinghua University \\ 
 \texttt{ \{yinyichun,shang.lifeng,jiang.xin,chen.xiao2,qun.liu\}@huawei.com} \\
 \texttt{c-chen19@mails.tsinghua.edu.cn} 
}

\date{}

\begin{document}
\maketitle
\renewcommand{\thefootnote}{\fnsymbol{footnote}}
\footnotetext[1]{Contribution during internship at Noah's Ark Lab.}
\renewcommand{\thefootnote}{\arabic{footnote}}

\begin{abstract}
Pre-trained language models (PLMs) have achieved great success in natural language processing. Most of PLMs follow the default setting of architecture hyper-parameters (e.g., the hidden dimension is a quarter of the intermediate dimension in feed-forward sub-networks) in BERT~\cite{Devlin2019BERTPO}. Few studies have been conducted to explore the design of architecture hyper-parameters in BERT, especially for the more {\it efficient} PLMs with tiny sizes, which are essential for practical deployment on resource-constrained devices. In this paper, we adopt the one-shot Neural Architecture Search (NAS) to automatically search architecture hyper-parameters. Specifically, we carefully design the techniques of one-shot learning and the search space to provide an adaptive and efficient development way of tiny PLMs for various latency constraints. We name our method AutoTinyBERT\footnote{Our code implementation and pre-trained models are available at \url{https://github.com/huawei-noah/Pretrained-Language-Model/tree/master/AutoTinyBERT}
.} and evaluate its effectiveness on the GLUE and SQuAD benchmarks. The extensive experiments show that our method outperforms both the SOTA search-based baseline (NAS-BERT) and the SOTA distillation-based methods (such as DistilBERT, TinyBERT, MiniLM and MobileBERT).  In addition, based on the obtained architectures, we propose a more efficient development method that is even faster than the development of a single PLM.
\end{abstract}
\section{Introduction}

\begin{figure}[t]
\centering
\includegraphics[width=0.5\linewidth]{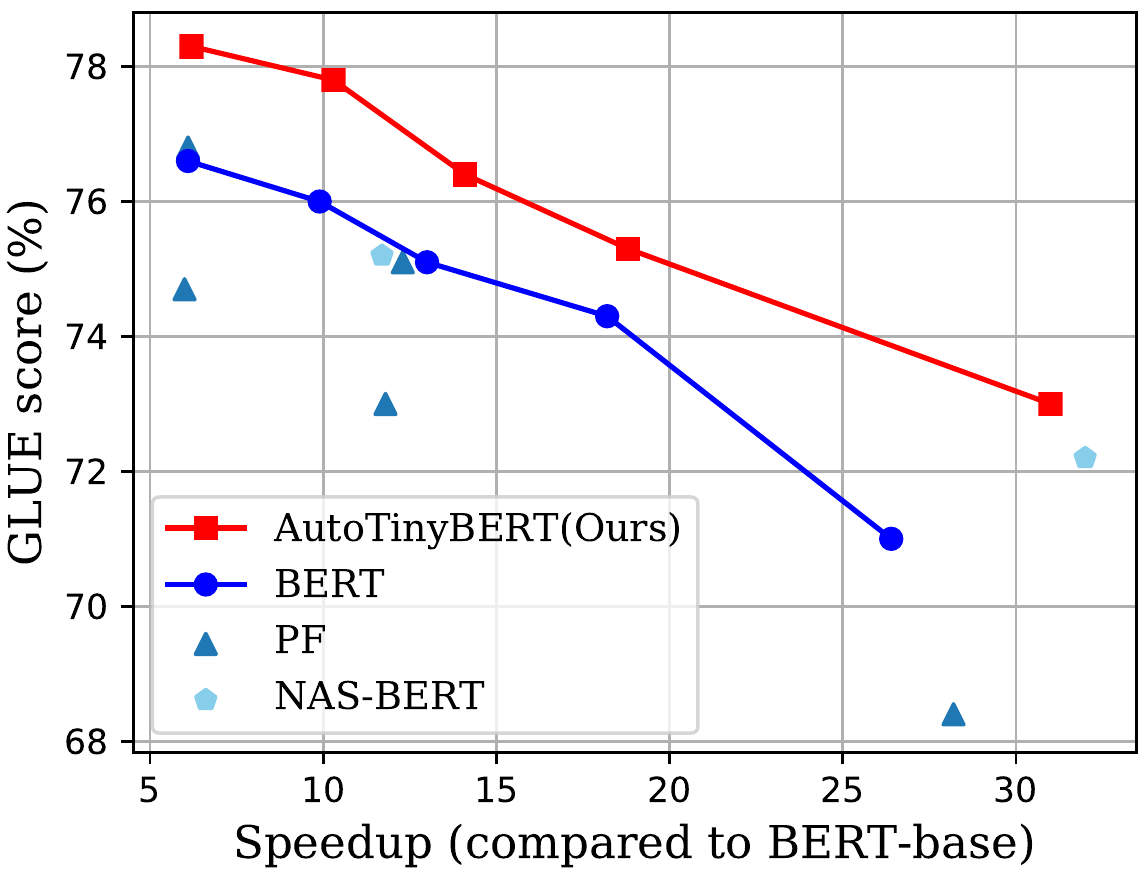}
\caption{Inference speedup vs. GLUE scores. Under the same speedup constraint, our method outperforms both the default hyper-parameter setting of BERT~\cite{Devlin2019BERTPO}, PF~\cite{turc2019well}) and  NAS-BERT~\cite{xu2021taskagnostic}. More details are in the Section~\ref{results_analysis}.}
\label{fig:perf}
\end{figure} 

Pre-trained language models, such as BERT~\cite{Devlin2019BERTPO}, RoBERTa~\cite{liu2019roberta} and XLNet~\cite{yang2019xlnet}, have become prevalent in natural language processing. To improve model performance, most PLMs (e.g. ELECTRA~\cite{clark2019electra} and GPT-2/3~\cite{radford2019language,brown2020language}) follow the default rule of hyper-parameter setting\footnote{The default rule is $d^{m}=d^{q|k|v}=1/4 d^{f}$, which means the dimension of hidden vector $d^{m}$ is equal to the dimensions of query/key/value vector $d^{q|k|v}$ and a quarter of the intermediate size $d^{f}$ in feed-forward networks.} in BERT to scale up their model sizes. Due to its simplicity, this rule has been widely used and can help large PLMs obtain promising results~\cite{brown2020language}.


In many industrial scenarios, we need to deploy PLMs on resource-constrained devices, such as smartphones and servers with limited computation power. Due to the expensive computation and slow inference speed, it is usually difficult to deploy PLMs such as BERT (12/24 layers, 110M/340M parameters) and GPT-2 (48 layers, 1.5B parameters) at their original scales. Therefore, there is an urgent need to develop PLMs with smaller sizes which have lower computation cost and inference latency. In this work, we focus on a specific type of {\it efficient } PLMs, which we define to have inference time less than 1/4 of BERT-base.\footnote{We empirically find that being at least 4x faster is a basic requirement in practical deployment environment.} 

Although, there have been quite a few work using knowledge distillation to build small PLMs~\cite{sanh2019distilbert,jiao2020tinybert,sun2019patient,sun12020mobilebert}, all of them focus on the application of distillation techniques~\cite{Hinton2015DistillingTK,romero2014fitnets} and do not study the effect of architecture hyper-parameter settings on model performance. Recently, neural architecture search and hyper-parameter optimization~\cite{tan2019efficientnet,han2020model} have been widely explored in machine learning, mostly in computer vision, and have been proven to find better designs than heuristic ones. Inspired by this research, one problem that naturally arises is \textit{can we find better settings of hyper-parameters\footnote{We abbreviate the phrase  {\it architecture hyper-parameter} as {\it hyper-parameter} in the paper.} for efficient PLMs?}

In this paper, we argue that the conventional hyper-parameter setting is not best for efficient PLMs (as shown in Figure~\ref{fig:perf}) and introduce a method to automatically search for the optimal hyper-parameters for specific latency constraints. Pre-training efficient PLMs is inevitably  resource-consuming~\cite{turc2019well}. Therefore, it is infeasible to directly evaluate millions of architectures. To tackle this challenge, we introduce the one-shot Neural Architecture Search (NAS)~\cite{brock2018smash, cai2018proxylessnas, yu2020bignas} to perform the automatic hyper-parameter optimization on efficient PLMs, named as AutoTinyBERT. Specifically, we first use the one-shot learning to obtain a big SuperPLM, which can act as {\it proxies} for all 
potential sub-architectures. {\it Proxy} means that when evaluating an architecture, we only need to extract the corresponding sub-model from the SuperPLM, instead of training the model from scratch. SuperPLM helps avoid the time-consuming pre-training process and makes the search process efficient. To make SuperPLM more effective, we propose practical techniques including the {\it head sub-matrix extraction} and {\it efficient batch-wise training}, and particularly limit the search space to the models with {\it identical layer structure}. Furthermore, by using SuperPLM, we leverage search algorithm~\cite{Xie2017ICCV, wang2020hat} to find hyper-parameters for various latency constraints.

In the experiments, in addition to the pre-training setting~\cite{Devlin2019BERTPO}, we also consider the setting of task-agnostic BERT distillation~\cite{sun12020mobilebert} that pre-trains with the loss of knowledge distillation, to build efficient PLMs.
Extensive results show that in pre-training setting, AutoTinyBERT not only consistently outperforms the BERT with conventional hyper-parameters under different latency constraints, but also outperforms NAS-BERT based on neural architecture search. In task-agnostic BERT distillation, AutoTinyBERT outperforms a series of existing SOTA methods of DistilBERT, TinyBERT and MobileBERT. 

Our contributions are three-fold: (1) we explore the problem of how to design hyper-parameters for efficient PLMs and introduce an effective and efficient method: AutoTinyBERT; (2) we conduct extensive experiments in both scenarios of pre-training and knowledge distillation, and the results show our method consistently outperforms baselines under different latency constraints; (3) we summarize a fast rule and it develops an AutoTinyBERT for a specific constraint with even about 50\% of the training time of a conventional PLM. 
\section{Preliminary}
\label{sec:preliminary}

\begin{figure*}[t]
\centering
\includegraphics[width=0.85\textwidth]{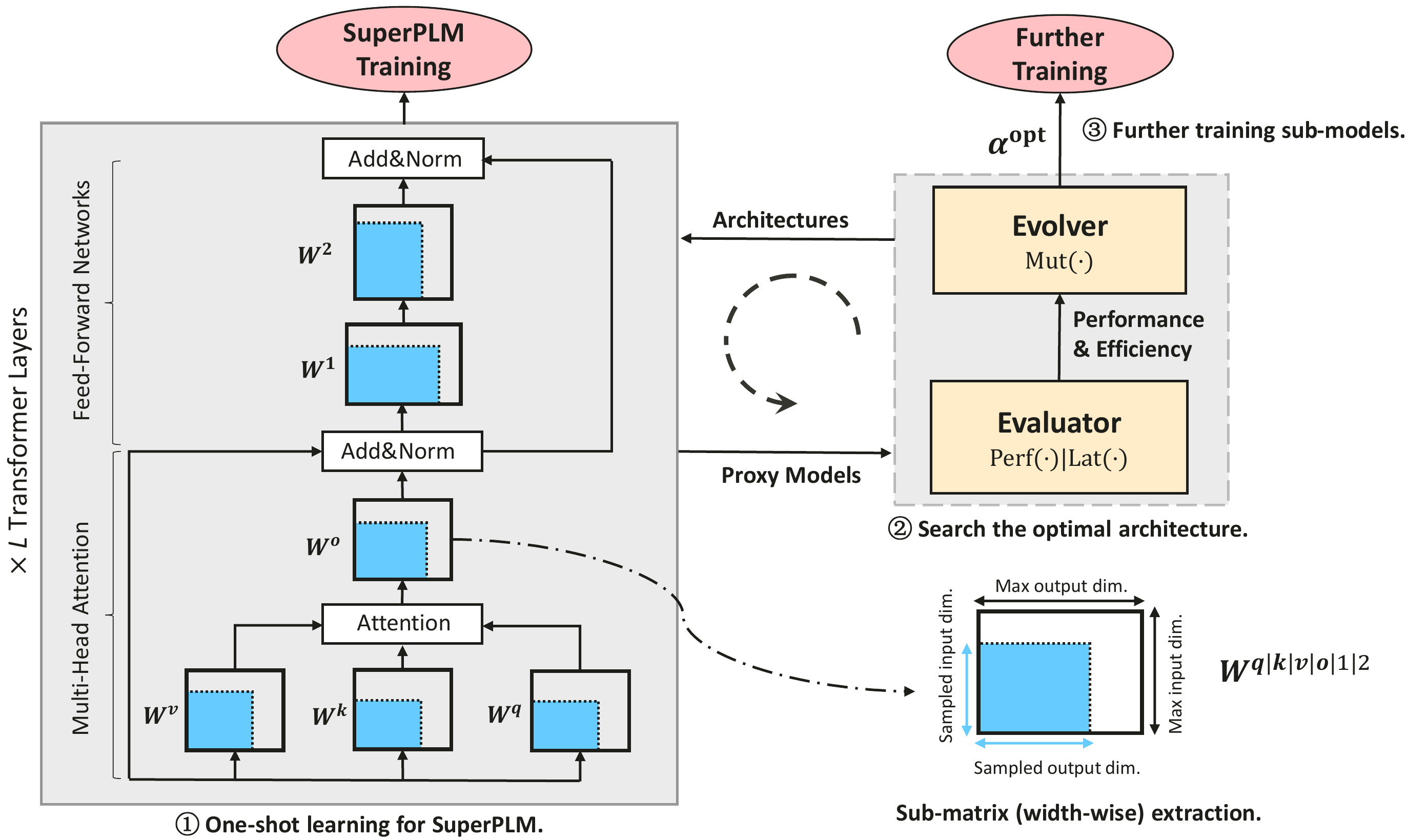}
\caption{Overview of AutoTinyBERT. We first train an effective SuperPLM with one-shot learning, where the objectives of pre-training or task-agnostic BERT distillation are used. Then, given a specific latency constraint, we perform an evolutionary algorithm on the SuperPLM to search optimal architectures. Finally, we extract the corresponding sub-models based on the optimal architectures and further train these models.}
\label{fig:overview_autobert}
\end{figure*}


Before presenting our method, we first provide some details about the Transformer layer~\cite{vaswani2017attention} to introduce the conventional hyper-parameter setting. Transformer layer includes two sub-structures: the multi-head attention (MHA) and the feed-forward network (FFN). 

For clarity, we show the MHA as a decomposable structure, where the MHA includes $h$ individual and parallel self-attention modules (called heads). The output of MHA is obtained by summing the output of all heads. Specifically, each head is represented by four main matrices  $\displaystyle \mW_i^q \in \displaystyle \R^{d^m\times d^q/h}$, $\displaystyle \mW_i^k \in \displaystyle \R^{d^m\times d^k/h}$, $\displaystyle \mW_i^v \in \displaystyle \R^{d^m\times d^v/h}$ and $\displaystyle \mW_i^o \in \displaystyle \R^{d^v/h\times d^o}$, and takes the hidden states\footnote{We omitted the batch size for simplicity.} $\displaystyle \mH \in \displaystyle \R^{l\times d^m}$ of the previous layer as input. The output of MHA is given by the following formulas: 
\begin{equation}
\begin{aligned}
   &{ \mQ_i, \mK_i, \mV_i} = {\mH\mW_i^{q}, \mH\mW_i^{k}, \mH\mW_i^{v} } \\
   &{\rm ATTN}(\mQ_i, \mK_i, \mV_i) = \softmax(\frac{\mQ_i {\mK_i}^{T}}{\sqrt{d^{q|k}/h}})\mV_i  \\
   &{\mH}_i = {\rm ATTN}(\mQ_i, \mK_i, \mV_i)\mW_i^o \\
   &{\rm MHA}(\mH) = \sum_{i=1}^h{\mH}_i,
\end{aligned}
\end{equation}
where $\mQ_i\in \displaystyle \R^{l \times d^q/h}$, $\mK_i \in \displaystyle \R^{l \times d^k/h} $, $\mV_i \in \displaystyle \R^{l \times d^v/h}$ are obtained by the linear transformations of $\displaystyle \mW_i^q$, $\displaystyle \mW_i^k$, $\displaystyle \mW_i^v$ respectively. $\rm{ATTN(\cdot)}$ is the scaled dot-product attention operation. Then output of each head is transformed to $\mH_{i}$ $\in \displaystyle \R^{l \times d^o} $ by $\mW_i^o$. Finally, outputs of all heads are summed as the output of MHA. In addition, residual connection and layer normalization are added on top of MHA to get the final output:
\begin{equation}
\mH^{\rm MHA}={\rm LayerNorm}(\mH + {\rm MHA}(\mH)).
\end{equation}
In the conventional setting of the hyper-parameters in BERT, all dimensions of matrices are the same as the dimension of the hidden vector, namely, $d^{q|k|v|o}$=$d^m$. In fact, there are only two requirements of $d^q$=$d^k$ and $d^o$=$d^m$ that must be satisfied because of the dot-product attention operation in MHA and the residual connection.

Transformer layer also contains an FFN that is stacked on the MHA, that is: 
\begin{equation}
{\mH^{\rm FFN}} = {\rm max}(0, \mH^{\rm MHA}\mW^1 + b_1)\mW^2 + b_2,
\end{equation}
where $\mW^1\in \displaystyle \R^{d^m \times d^f}$, $\mW^2\in \displaystyle \R^{d^f \times d^m}$, $b_1 \in \displaystyle \R^{d^f}$ and $b_2 \in \displaystyle \R^{d^m}$. Similarly, there are modules of residual connection and layer normalization on top of FFN. In the original Transformer, $d^f$=$4d^m$ is assumed. Thus, we conclude that the conventional hyper-parameter setting follows the rule of \{$d^{q|k|v|o}$=$d^m$, $d^f$=$4d^m$\}.

\section{Methodology}
\subsection{Problem Statement}
Given a constraint of inference time, our goal is to find an optimal configuration of architecture hyper-parameters $\alpha^{\rm opt}$ built with which PLM can achieve the best performances on downstream tasks. This optimization problem is formulated as:
\begin{equation}
\begin{aligned}
\alpha^{\rm opt} = &\argmax_{\alpha \in \displaystyle \gA} \rm{Perf}(\alpha, \theta^*_{\alpha}), \\
\textrm{s.t.}\ \ \ \ \theta^*_{\alpha} = &\argmin_{\theta} L_{\alpha}(\theta), \ {\rm Lat}(\alpha)  \leq T,\\
\end{aligned}
\end{equation}
where $T$ is a specific time constraint, $\displaystyle \gA$ refers to the set of all possible architectures (i.e., combination of hyper-parameters), $\rm{Lat}(\cdot)$ is a latency evaluator, $L_{\alpha}(\cdot)$ denotes the loss function of PLMs with the hyper-parameter $\alpha$, and $\theta$ is the corresponding model parameters. We aim to search an optimal architecture for efficient PLM (${\rm Lat}(\alpha)<1/4\times{\rm Lat}({\rm BERT_{base}})$).


\subsection{Overview}

A straightforward way to get the optimal architecture is to enumerate all possible architectures. However, it is infeasible because each trial involves a time-consuming pre-training process. Therefore, we introduce one-shot NAS to search $\alpha^{\rm opt}$, as shown in the Figure~\ref{fig:overview_autobert}. The proposed method includes three stages: (1) the one-shot learning to obtain SuperPLM that can be used as the proxy for various architectures; (2) the search process for the optimal hyper-parameters; (3) the further training with the optimal architectures and corresponding sub-models. In the following sections, we first introduce the search space, which is the basis for the one-shot learning and search process. Then we present the three stages respectively.

\begin{figure}[t]
\centering
\includegraphics[width=0.48\textwidth]{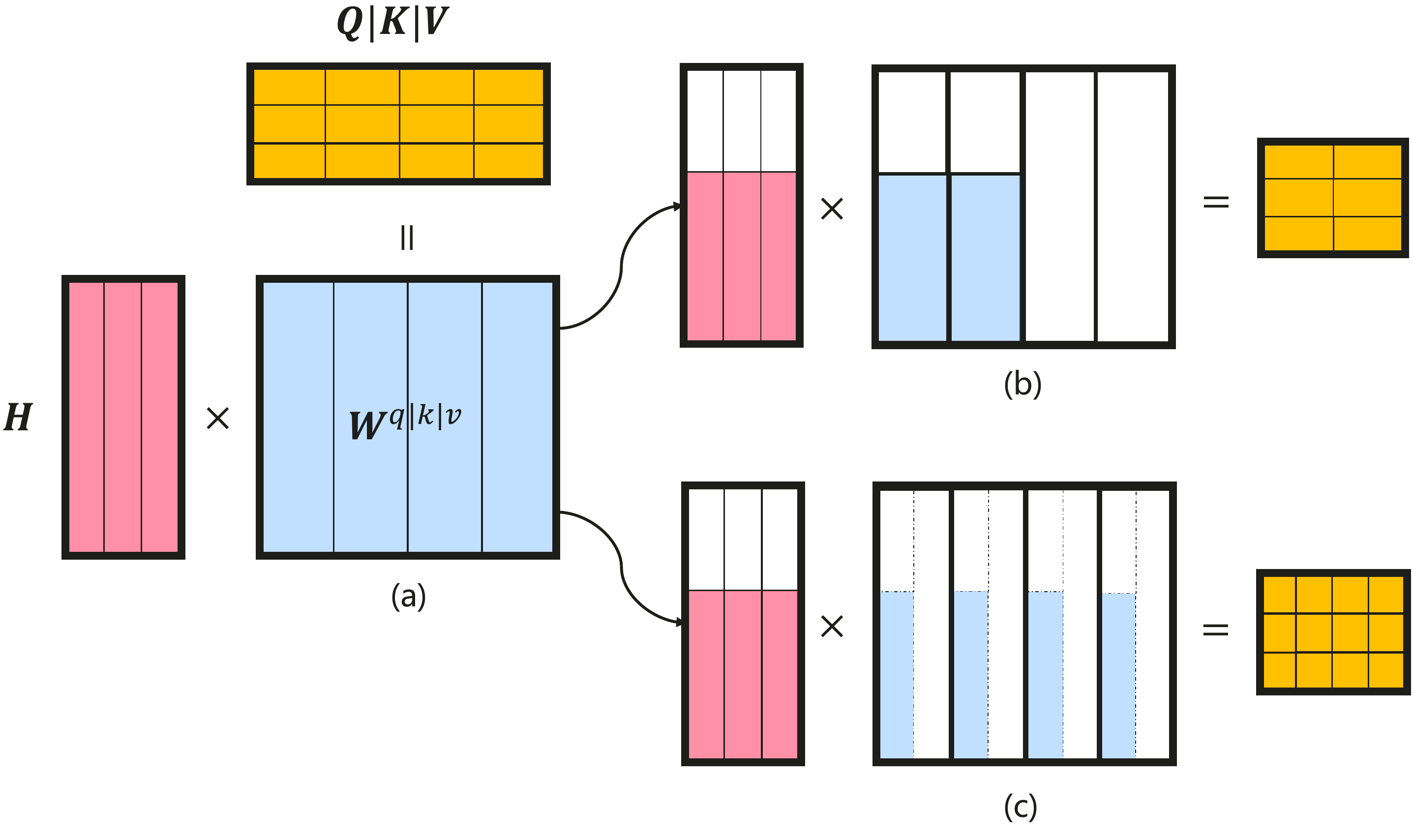}
\caption{MHA sub-matrix extraction. (a) means that the original matrix operation where we take four heads and three hidden vectors as an example. White boxes refer to the un-extracted parameters. (b) means that we extract heads while keeping the dimension per head. (c) means that we extract parameters from each head while keeping the head number as the original matrix.}
\label{fig:sub-matrix}
\end{figure}

\subsection{Search Space}
From the Section~\ref{sec:preliminary}, we know that the conventional hyper-parameter setting is: \{$d^{q|k|v|o}$=$d^m$, $d^f$=$4d^m$\}, which is widely-used in PLMs. The architecture of a PLM is parameterized as:
$\alpha = \{l^t, d^m, d^q, d^k, d^v, d^f, d^o\}$, which is subjected to the constraints \{$d^q=d^k, d^o=d^m$\}. Let $l^t$ denote the layer number and $d^{*}$ refer to different dimensions in the Transformer layer. We denote the search space of $l^t$ and $d^{*}$ as $\displaystyle \gA_{l^t}$ and $\displaystyle \gA_{d^*}$ respectively. The overall search space is: $\displaystyle \gA  = \gA_{l^t} \times \gA_{d^{m|o}} \times \gA_{d^{q|k}} \times \gA_{d^v} \times  \gA_{d^f}$.

In this work, we only consider the case of identical structure for each Transformer layer, instead of the non-identical Transformer~\cite{wang2020hat} or other heterogeneous modules~\cite{xu2021taskagnostic} (such as convolution units). It has two advantages: (1) it reduces an exponential search space of $\mathcal{O}(\prod \limits_{*} \displaystyle |\gA_{d^*}| ^{|\gA_{l^t}|})$ to a linear search space of $\mathcal{O}(\prod \limits_{*} \displaystyle |\gA_{d^*}| |\gA_{l^t}|)$, greatly reducing the number of possible architectures in SuperPLM training and the exploration space in the search process. It leads to a more efficient search process. (2) An identical and homogeneous structure is in fact more friendly to hardware and software frameworks, e.g., Hugging Face Transformer~\cite{wolf-etal-2020-transformers}. With a few changes, we can use the original code to use AutoTinyBERT, as shown in Appendix~\ref{appendix:code}.

\renewcommand{\algorithmicrequire}{\textbf{Input:}} 
\renewcommand{\algorithmicensure}{\textbf{Output:}} 

 \begin{algorithm}[t] 
  \caption{Batch-wise training for SuperPLM} 
	 \label{alg:batch_train}
  \begin{algorithmic}[1] 
   \Require All possible candidates $\displaystyle \gA$; Training thread (GPU) number $N$; Large-scale unsupervised dataset $D$; Training epochs $E$. Sample times $M$ per batch. SuperPLM parameters ($\theta$).
   \Ensure Trained SuperPLM ($\theta$)
    \For{$t = 1 \to E$}
				\For {$batch$ in $D$}
						\State Divide $batch$ into $N$ $sub\_batches$
						\State Distribute $sub\_batches$ to $N$ threads
						\State Clear the gradients
						\For{$m = 1 \to M$}
								\State Sample $N$ sub-models from $\displaystyle \gA$ 
								\State Distribute sub-models to threads
								\State Calculate gradients in each thread
						\EndFor \\
\ \ \ \ \ \ \ \ \ \ \ \ Update the $\theta$ with the average gradients
				\EndFor
	\EndFor
			
  \end{algorithmic} 
 \end{algorithm}

\subsection{One-shot Learning for SuperPLM}

We employ the one-shot learning~\cite{brock2018smash, yu2020bignas} to obtain a SuperPLM whose sub-models can act as the proxy for PLMs trained from scratch. The configurations of SuperPLM in this work are $l^t$=$8$, $d^{m|q|k|v|o}$=$768$, and $d^f$=$3072$. In each step of the one-shot learning, we train several sub-models randomly sampled from SuperPLM to make their performance close to the models trained from scratch. Although the sampling/search space has been reduced to linear complexity, there are still more than 10M possible sub-structures in SuperPLM (the details are shown in the Appendix~\ref{appendix:space}). Therefore, we introduce an effective batch-wise training method to cover the sub-models as much as possible. Specifically, in parallel training, we first divide each batch into multiple sub-batches and distribute them to different threads as parallel training data. Then, we sample several sub-models on each thread for training and merge the gradients of all threads to update the SuperPLM parameters. We illustrate the training process in the Algorithm~\ref{alg:batch_train}.

\begin{table*}[]
	\centering
	\scalebox{0.72}{
		\begin{tabular}{l|c|ccccccccc|cc}
			\toprule
			Model & Speedup & SQuAD & SST-2 & MNLI & MRPC & CoLA & QNLI & QQP & STS-B & RTE & Score &  Avg.\\ \midrule
{\bf AutoTinyBERT-S1} & 7.2$\times$   & 83.3 & 89.4 & 79.4 & 85.5 & 42.4 & 87.3 & 88.8 & 87.5 & 66.3 & \textbf{78.3} & \textbf{78.9} \\ 
BERT-S1    & 7.1$\times$ & 81.5 & 88.9 & 78.4 & 81.3 & 35.8 & 86.4 & 88.2 & 86.7 & 66.4 & 76.5 & 77.1 \\
PF-4L512D\ddag~\cite{turc2019well} & 7.1$\times$ & 81.7    &  89.4    &   78.5   &   82.8   & 35.2 &  87.0   &  88.6    & 87.4      &    65.7     & 76.8 & 77.4 \\
PF-2L768D\ddag~\cite{turc2019well} & 7.0$\times$  & 71.1  & 88.8  &  76.5 &   79.6 &  26.7 &  84.9    & 88.1     &   86.6 & 67.1      & 74.8 & 74.4 \\ \midrule 
{\bf AutoTinyBERT-S2} & 15.7$\times$ & 78.1 & 88.2 & 76.8 & 82.8 & 35.5 & 85.4 & 87.8 & 86.5 & 68.2 & \textbf{76.4} & \textbf{76.6} \\
BERT-S2      & 14.8$\times$ & 77.6 & 87.5 & 76.5 & 79.6 & 32.8 & 84.4 & 87.0 & 86.6 & 66.4 & 75.1 & 75.4 \\
NAS-BERT$_{10}\dag$~\cite{xu2021taskagnostic} & 12.7$\times$ & -    & 88.6 & 76.0 & 81.5 & 27.8 & 86.3 & 88.4 & 84.3 & 68.7 & 75.2 & - \\ 
PF-2L512D\ddag~\cite{turc2019well} & 12.8$\times$ & 69.2 & 87.1 & 74.7 & 76.9 & 23.2 & 84.4 & 87.0 & 86.0 & 64.9 & 73.0 & 72.6 \\ 
PF-6L256D\ddag~\cite{turc2019well} & 13.3$\times$ &   77.0   &   87.6   &   76.4   &  80.3  &  33.2 &  85.7 & 86.7  &  86.0  & 64.9  & 75.1 & 75.3  \\ \midrule
{\bf AutoTinyBERT-S3} & 20.2$\times$ & 75.8 & 86.8 & 76.4 & 80.4 & 33.2 & 85.0 & 87.6 & 86.7 & 66.4 & \textbf{75.3} & \textbf{75.4} \\
BERT-S3           & 20.1$\times$ & 73.7 & 86.4 & 75.0 & 81.3 & 31.2 & 84.0 & 87.1 & 85.8 & 63.8 & 74.3 & 74.3 \\ \midrule
{\bf AutoTinyBERT-S4} & 31.0$\times$ & 71.9	& 86.5 & 74.2 & 81.9 &	17.6 & 84.6 & 86.5 & 85.9 &	66.7 & \textbf{73.0} & \textbf{72.9} \\ 
BERT-S4     & 31.3$\times$ & 69.5 & 85.5 & 73.9 & 76.9 & 15.9 & 83.9 & 85.9 & 85.3 & 61.0 & 71.0 & 70.9 \\
NAS-BERT$_{5}\dag$~\cite{xu2021taskagnostic} &  32.0$\times$ & -  & 84.9 & 74.2 & 80.0 & 19.6 & 83.9 & 85.7 & 82.8 & 67.0 & 72.3 & - \\ 
PF-6L128D\ddag~\cite{turc2019well} & 28.2$\times$ & 63.6  & 84.6  & 72.3 & 78.6 & 0  & 83.3  & 83.8 & 84.5 & 65.7 & 69.1  & 68.5 \\ \midrule 
\end{tabular}}
\caption{Comparison between AutoTinyBERT and baselines in pre-training setting. The results are evaluated on the dev set of GLUE benchmark and SQuADv1.1. We use the metric of Matthews correlation for CoLA, F1 for SQuADv1.1, Pearson-Spearman correlation for STS-B, and accuracy for other tasks. We report the average score excluding SQuAD (Score) in addition to the average score of all tasks (Avg.). The speedup is in terms of the BERT$_{\rm base}$ inference speed and evaluated on a single CPU with a single input of 128 length. PF-$\rm x$L$\rm y$D, the $\rm x$ and $\rm y$ refer to the layer number and hidden dimension respectively. \dag denotes that the results are taken from~\cite{xu2021taskagnostic} and \ddag denotes that the results are obtained by fine-tuning the released models.}
\label{tab:pretrain_results}
\end{table*}

Given a specific hyper-parameter setting $\alpha = \{l^t, d^m, d^q, d^k, d^v, d^f, d^o\}$, we get a sub-model from SuperPLM by the depth-wise and width-wise extraction. Specifically, we first perform the depth-wise extraction that extracts the first $l^t$ Transformer layers from SuperPLM, and then perform the width-wise extraction that extracts bottom-left sub-matrices from original matrices. For MHA, we apply two strategies illustrated in Figure~\ref{fig:sub-matrix} : (1) keep the dimension of each head same as SuperPLM, and extract some of the heads; (2) keep the head number same as SuperPLM, and extract sub-dimensions from each head. The first strategy is the standard one and we use it for pre-training and the second strategy is used for task-agnostic distillation because that attention-based distillation~\cite{jiao2020tinybert} requires the student model to have the same head number as the teacher model.

\subsection{Search Process}
In the search process, we adopt an evolutionary algorithm~\cite{Xie2017ICCV,jiao2020improving}, where Evolver and Evaluator interact with each other to evolve better architectures. Our search process is efficient, as shown in the Section~\ref{sec:fast_rule}.

Specifically, Evolver firstly samples a generation of architectures from $\displaystyle \gA$. Then Evaluator extracts the corresponding sub-models from SuperPLM and ranks them based on their performance on tasks of SQuAD and MNLI. The architectures with the high performance are chosen as the winning architectures and Evolver performs the mutation $\rm Mut(\cdot)$ operation on the winning ones to produce a new generation of architectures. This process is conducted repeatedly. Finally, we choose several architectures with the best performance for further training. We use $\rm{Lat}(\cdot)$ to predict the latency of the candidates to filter out the candidates that do not meet the latency constraint. $\rm{Lat}(\cdot)$ is built with the method by~\citet{wang2020hat}, which first samples about 10k architectures from $\displaystyle \gA$ and collects their inference time on target devices, and then uses a feed-forward network to fit the data. For more details of evolutionary algorithm, please refer to Appendix~\ref{appendix:evo_alg}. Note that we can use different methods in search process, such as random search and more advanced search, which is left as future work.

\subsection{Further Training} 
The search process produces top several architectures, with which we extract these corresponding sub-models from SuperPLM and continue training them using the pre-training or KD objectives.

\section{Experiment}

\begin{table*}[]
	\centering
	\scalebox{0.66}{
		\begin{tabular}{l|c|ccccccccc|cc}
			\toprule
			Model & Speedup & SQuAD & SST-2 & MNLI & MRPC$\mathparagraph$ & CoLA & QNLI & QQP$\mathparagraph$ & STS-B & RTE & Score & Avg. \\ \midrule 
\multicolumn{8}{l}{\it Dev results on GLUE and dev result on SQuAD} \\ \midrule
{\bf AutoTinyBERT-KD-S1} & 4.6$\times$ & 87.6 & 91.4 & 82.3 & 88.5 & 47.3 & 89.7 & 89.9 & 89.0 & 71.1 & \textbf{81.2} & \textbf{81.9}  \\ 
{BERT-KD-S1}    & 4.9$\times$ & 86.2 & 89.7 & 81.1 & 87.9 & 41.8 & 87.3 & 88.4 & 88.4 & 68.2 & 79.1 & 79.9 \\
MobileBERT$_{\rm Tiny}$\ddag\cite{sun12020mobilebert} & 3.6*$\times$          & 88.6 & 91.6 & 82.0 & 86.7 &   -   &  -   &  -   &   -  &  -   &   -   &  -    \\ \midrule
{\bf AutoTinyBERT-KD-S2} & 9.0$\times$ & 84.6 & 88.8 & 79.4 & 87.3 & 32.2 & 88.0 & 87.7 & 88.0 & 68.9 & \textbf{77.5} & \textbf{78.3} \\ 
BERT-KD-S2      & 9.8$\times$       & 82.5 & 87.8 & 77.9 & 86.5 & 31.5 & 86.9 & 87.6 & 87.4 & 66.4 & 76.5 & 77.2 \\  
MiniLM-4L312D\dag~\cite{wang2020minilm} & 9.8$\times$       & 82.1 & 87.3 & 78.3 & 83.6 & 26.3 & 87.1 & 87.3 & 86.3 & 62.4 & 74.8 & 75.6 \\ 
TinyBERT-4L312D\dag $\mathsection$~\cite{jiao2020tinybert} & 9.8$\times$ & 81.0 & 87.8 & 76.9 & 77.9 & 22.9 & 86.0 & 87.7 & 83.3 & 58.8 & 72.7 & 73.6 \\
\midrule
{\bf AutoTinyBERT-KD-S3} & 10.7$\times$ & 83.3 & 88.3 & 78.2 & 85.8 & 29.1 & 87.4 & 87.4 & 86.7 & 66.4 & \textbf{76.2} & \textbf{77.0}\\
BERT-KD-S3            & 11.7$\times$ & 81.6 & 86.5 & 76.8 & 82.5 & 27.6 & 85.6 & 86.5 & 86.2 & 64.9 & 74.6 & 75.4 \\ \midrule
{\bf AutoTinyBERT-KD-S4} & 17.0$\times$  & 78.7 & 86.8 & 76.0 & 81.4 & 20.4 & 85.5 & 86.9 & 86.0 & 64.9 & \textbf{73.5} & \textbf{74.1} \\ 
BERT-KD-S4       & 17.0$\times$ & 77.4 & 85.7 & 75.4 & 80.3 & 18.9 & 85.0 & 85.9 & 84.7 & 63.1 & 72.4 & 72.9 \\ \midrule
\multicolumn{8}{l}{\it Test results on GLUE and dev result on SQuAD} \\ \midrule
{\bf AutoTinyBERT-KD-S1} & 4.6$\times$    & 87.6 & 90.6 & 81.2 & 88.9 & 44.7 & 87.4 & 70.5 & 85.1 & 64.8 & \textbf{76.7} & \textbf{77.9} \\ 
BERT-3L-PKD\ddag~\cite{sun2019patient}  &  4.1$\times$ & -    & 87.5 & 76.7 & 80.7 & -    & 84.7 & 68.1 & -    & 58.2 & - & - \\
DistilBERT-4L\ddag~\cite{sanh2019distilbert} &         3.0$\times$ & 81.2 & 91.4 & 78.9 & 82.4 & 32.8 & 85.2 & 68.5 & 76.1 & 54.1 & 71.2 & 72.3 \\
TinyBERT-4L516D\dag$\mathsection$~\cite{jiao2020tinybert}  &4.9$\times$ & 84.6 & 88.2 & 80.0 & 86.3 & 27.9 & 85.6 & 69.1 & 83.0 &  61.5& 72.7 & 74.0 \\
MiniLM-4L516D\dag~\cite{wang2020minilm}     &     4.9$\times$           & 85.5 & 90.0 & 80.2 & 87.2 & 39.1 & 86.5 & 70.0 & 83.4 &  63.7& 75.0 & 76.2 \\ 
MobileBERT$_{\rm Tiny}\ddag$~\cite{sun12020mobilebert} &3.6*$\times$    & 88.6 & 91.7 & 81.5 & 87.9 & 46.7 & 89.5 & 68.9 & 80.1 & 65.1  & 76.4 & 77.8 \\ \midrule
\end{tabular}}
\caption{Comparison between AutoTinyBERT and baselines based on knowledge distillation. $\ddag$ denotes that the results are taken from \cite{sun12020mobilebert} and $\dag$ means the models trained using the released code or the re-implemented code with ELECTRA$_{\rm base}$ as the teacher model. $\mathparagraph$ means these tasks use accuracy for dev set and F1 for test set respectively. $\mathsection$ denotes the task-agnostic TinyBERT without task-specific distillation. * means that the speedup is different from the ~\cite{sun12020mobilebert}, because it is evaluated on a Pixel phone and not on server CPUs. - means that the results are missing in the original paper. Other information refer to the Table~\ref{tab:pretrain_results}.}
\label{tab:kd_results}
\end{table*}

\subsection{Experimental Setup}
{\bf Dataset and Fine tuning.}
We conduct the experiments on the GLUE benchmark~\cite{wang2018glue} and SQuADv1.1~\cite{rajpurkar2016squad}. For GLUE, we set the batch size to 32, choose the learning rate from \{1e-5, 2e-5, 3e-5\} and choose the epoch number from \{4, 5, 10\}. For SQuADv1.1, we set the batch size to 16, the learning rate to 3e-5 and the epoch number to 4. The details for all datasets are displayed in Appendix~\ref{appendix:hyper-parameter}.

\vspace{0.3cm}

\noindent {\bf AutoTinyBERT.} 
Both the one-shot and further training use BooksCorpus~\cite{zhu2015aligning} and English Wikipedia as training data. The settings for one-shot training are: peak learning rate of 1e-5, warmup rate of 0.1, batch size of 256 and 5 running epochs. Further training follows the same setting as the one-shot training except for the warmup rate of 0. In the batch-wise training algorithm~\ref{alg:batch_train}, the thread number $N$ is set to 16, the sample times $M$ per batch is set to 3, and epoch number $E$ is set to 5. We train the SuperPLM with an architecture of \{$l^t$=$8$, $d^{m|q|k|v|o}$=$768$, $d^f$=$3072$\}. In the search process, Evolver performs 4 iterations with a population size of 25 and it chooses top three architectures for further training. For more details of the sampling/search space and evolutionary algorithm, please refer to Appendix~\ref{appendix:space} and~\ref{appendix:evo_alg}.

We train AutoTinyBERT in both ways of pre-training~\cite{Devlin2019BERTPO} and task-agnostic BERT distillation~\cite{sun12020mobilebert}. For task-agnostic distillation, we follow the first stage of TinyBERT~\cite{jiao2020tinybert} except that only the last-layer loss is used, and ELECTRA$_{\rm base}$~\cite{clark2019electra} is used as the teacher model.

\noindent {\bf Baselines.} For the pre-training baselines, we include PF ({\it Pre-training + Fine-tuning}, proposed by ~\citet{turc2019well}), BERT-S* (BERT under several hyper-parameter configurations), and NAS-BERT~\cite{xu2021taskagnostic}. Both PF and BERT-S* follow the conventional setting rule of hyper-parameters. BERT-S* uses the training setting: peak learning rate of 1e-5, warmup rate of 0.1, batch size of 256 and 10 running epochs. NAS-BERT searches the architecture built on the non-identical layer and heterogeneous modules. For the distillation baselines, we compare some typical methods, including DistilBERT, BERT-PKD, TinyBERT, MiniLM, and MobileBERT. The first four methods use the conventional architectures. MobileBERT is equipped with a bottleneck structure and a carefully designed balance between MHA and FFN. We also consider BERT-KD-S*, which use the same training setting of BERT-S*, except for the loss of knowledge distillation. BERT-KD-S* also uses ELECTRA$_{\rm base}$ as the teacher model.

\subsection{Results and Analysis} 
\label{results_analysis}
The experiment is conducted under different latency constraints that are from 4$\times$ to 30$\times$ faster than the inference of BERT$_{\rm base}$. The results of pre-training and task-agnostic distillation are shown in the Table~\ref{tab:pretrain_results} and Table~\ref{tab:kd_results} respectively. 

\begin{figure*}[t]
\centering
\includegraphics[width=0.9\textwidth]{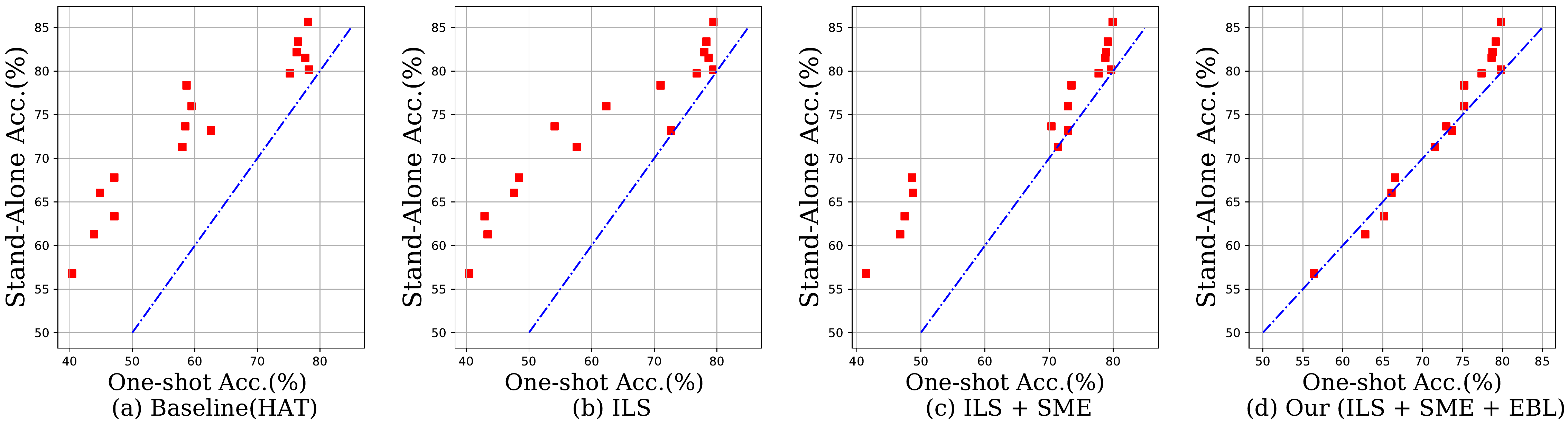}
\caption{Ablation study of one-shot SuperPLM learning.  Acc. means the average score on SQuAD and MNLI. The dashed line represents the function of y=x.}
\label{fig:SuperPLM_study}
\end{figure*}

We observe that in the settings of the pre-training and knowledge distillation, the performance gap of different models with similar inference time is obvious, which shows the necessity of architecture optimization for efficient PLMs. In the Table~\ref{tab:pretrain_results}, some observations are: (1) the architecture optimization methods of AutoTinyBERT and NAS-BERT outperform both BERT and PF that use the default architecture hyper-parameters; (2) our method outperforms NAS-BERT that is built with the non-identical layer and heterogeneous modules, which shows that the proposed method is effective for the architecture search of efficient PLMs. In the Table~\ref{tab:kd_results}, we observe that: (1) our method consistently outperforms the conventional structure in all the speedup constraints; (2) our method outperforms the classical distillation methods (e.g., BERT-PKD, DistilBERT, TinyBERT, and MiniLM) that use the conventional architecture. Moreover, AutoTinyBERT achieves comparable results with MobileBERT, and its inference speed is 1.5$\times$ faster. 


\begin{table}[]	
	\centering
	\scalebox{0.7}{
		\begin{tabular}{l|cc|c}
			\toprule
Training  & SQuAD & MNLI & Pairwise \\
Method  &  F1(\%)  & Acc.(\%) & Acc.(\%) \\ \midrule
Stand-alone & 71.2 & 76.3 & 100 \\ \midrule
Baseline & 50.1 & 72.7 & 90.0 \\ 
ILS & 52.9 & 73.4 & 91.6 \\ 
ILS + SME & 59.5 & 74.1 & 94.2 \\ 
ILS + SME + EBL (Ours) & {\bf 70.5} & {\bf 74.4} & {\bf 96.7} \\ \midrule
\end{tabular}}
\caption{Ablation Study of SuperPLM. ILS, SME and EBL mean that the identical layer structure, MHA sub-matrix extraction and effective batch-wise training.} \label{tab:ablation_results}
\end{table}

\subsection{Ablation Study of One-shot Learning} 
We demonstrate the effectiveness of one-shot learning by comparing the performance of one-shot model and stand-alone trained model on the given architectures. We choose 16 architectures and their corresponding PF models\footnote{The first 16 models \url{https://github.com/google-research/bert} from 2L128D to 8L768D.} as the evaluation benchmark. The pairwise accuracy is used as a metric to indicate the ranking correction between the architectures under one-shot training and the ones under stand-alone full training~\cite{luo2019balanced} and its formula is described in Appendix ~\ref{appendix:pair_acc}.

We do the ablation study to analyze the effect of proposed identical layer structure (ILS), MHA sub-matrix extraction (SME) and effective batch-wise learning (EBL) on SuperPLM learning. Moreover, we introduce HAT~\cite{wang2020hat}, as a baseline of one-shot learning. HAT focuses on the search space of non-identical layer structures. The results are displayed in Table~\ref{tab:ablation_results} and Figure~\ref{fig:SuperPLM_study}. 

\begin{table}[]
	\centering
	\scalebox{0.7}{
		\begin{tabular}{l|l|l|c}
		\toprule
		  Version & BERT  & AutoTinyBERT     & Speedup \\ \midrule  
		  \multicolumn{4}{l}{\it Pre-training}    \\ \midrule
           S1 & 4-512-2048-8-512 & 5-564-1054-8-512 & 7.1/7.2$\times$ \\
		  S2 & 4-320-1280-5-320 & 4-396-624-6-384 & 14.8/15.7$\times$ \\
		  S3 & 4-256-1024-4-256 & 4-432-384-4-256 & 20.1/20.2$\times$ \\
           S4 & 4-192-768-3-192 & 3-320-608-4-256 & 28.4/27.2$\times$ \\  \midrule
		  \multicolumn{4}{l}{\it Task-agnostic BERT Distillation}    \\ \midrule
		  KD-S1 & 4-512-2048-12-516 & 5-564-1024-12-528 & 4.9/4.6$\times$ \\
		  KD-S2\dag & 4-312-1200-12-312 & 5-324-600-12-324 & 9.8/9.0$\times$ \\ 
		  KD-S3 & 4-264-1056-12-264 & 5-280-512-12-276 & 11.7/10.7$\times$ \\
	      KD-S4 & 4-192-768-12-192 & 4-256-480-12-192 &  17.0/17.0$\times$ \\ \midrule
		\end{tabular}}
\caption{BERT and AutoTinyBERT architectures under the different speedup constraints. The architecture is formatted as ``$l^t$-$d^{m|o}$-$d^f$-$h$-$d^{q|k|v}$". We assume that $d^{q|k}=d^v$ in the experiment for the training and search efficiency. $\dag$ means that we use the structure of TinyBERT and do not strictly follow the conventional rule.}
\label{tab:arches}
\end{table}

It can be seen from the figure that compared with stand-alone trained models, the HAT baseline has a significant performance gap, especially in small sizes. Both ILS and SME benefit the one-shot learning for large and medium-sized models. When further combined with EBL, SuperPLM can obtain similar or even better results than stand-alone trained models of small sizes and perform close to stand-alone trained models of big sizes. The results of the table show that: (1) the proposed techniques have positive effects on SuperPLM learning, and EBL brings a significant improvement on a challenging task of SQuAD; (2) SuperPLM achieves a high pairwise accuracy of 96.7\% which indicates that the proposed SuperPLM can be a good proxy model for the search process; (3) the performance of SuperPLM is still a little worse than the stand-alone trained model and we need to do the further training to boost the performance.


\begin{figure*}[t]
\begin{minipage}{.53\linewidth}
	\centering
	\scalebox{0.70}{
\begin{tabular}{lcccc}
\toprule
\multicolumn{1}{l|}{\multirow{2}{*}{Method}} & \multicolumn{1}{c|}{\multirow{2}{*}{Speedup}} & \multicolumn{1}{c|}{Search Cost}              & \multicolumn{1}{c|}{Training Cost}          & \multirow{2}{*}{Avg.} \\
\multicolumn{1}{l|}{}                        &     \multicolumn{1}{l|}{}                                           & \multicolumn{1}{c|}{(GPU Hours)}     & \multicolumn{1}{c|}{(GPU Hours)} &                        \\ \midrule
                   \multicolumn{1}{l|}{BERT-S5}  &   \multicolumn{1}{c|}{9.9$\times$}         &      \multicolumn{1}{c|}{\bf 0}     & \multicolumn{1}{c|}{580}              &         76.1               \\ 
                   \multicolumn{1}{l|}{\bf AutoTinyBERT-S5} &     \multicolumn{1}{c|}{\bf 10.8$\times$}     &      \multicolumn{1}{c|}{150}   &  \multicolumn{1}{c|}{870}             &             {\bf 77.9}           \\
				\multicolumn{1}{l|}{\bf AutoTinyBERT-Fast-S5} &      \multicolumn{1}{c|}{ 10.3$\times$}    &        \multicolumn{1}{c|}{12}     &     \multicolumn{1}{c|}{\bf 290}      &                77.6       \\   \midrule 
\end{tabular}}
\captionof{table}{Computation cost of different methods. AutoTinyBERT and AutoTinyBERT-Fast have 100 and 8 architectures ($\times$1.5 V100 GPU hour) respectively to be tested in the search process. AutoTinyBERT performs further 5 epochs ($\times$58 V100 GPU hours) training for top three architectures, BERT is trained from scratch with 10 epochs, and AutoTinyBERT-Fast does the further training for one architecture. We give more information including the model architectures and detailed scores of all tasks in the Appendix~\ref{appendix:details_fast}.}
\label{tab:10x_speedup} 
\end{minipage}
\hfill
\begin{minipage}{.43\linewidth}
\centering
\includegraphics[width=0.7\textwidth]{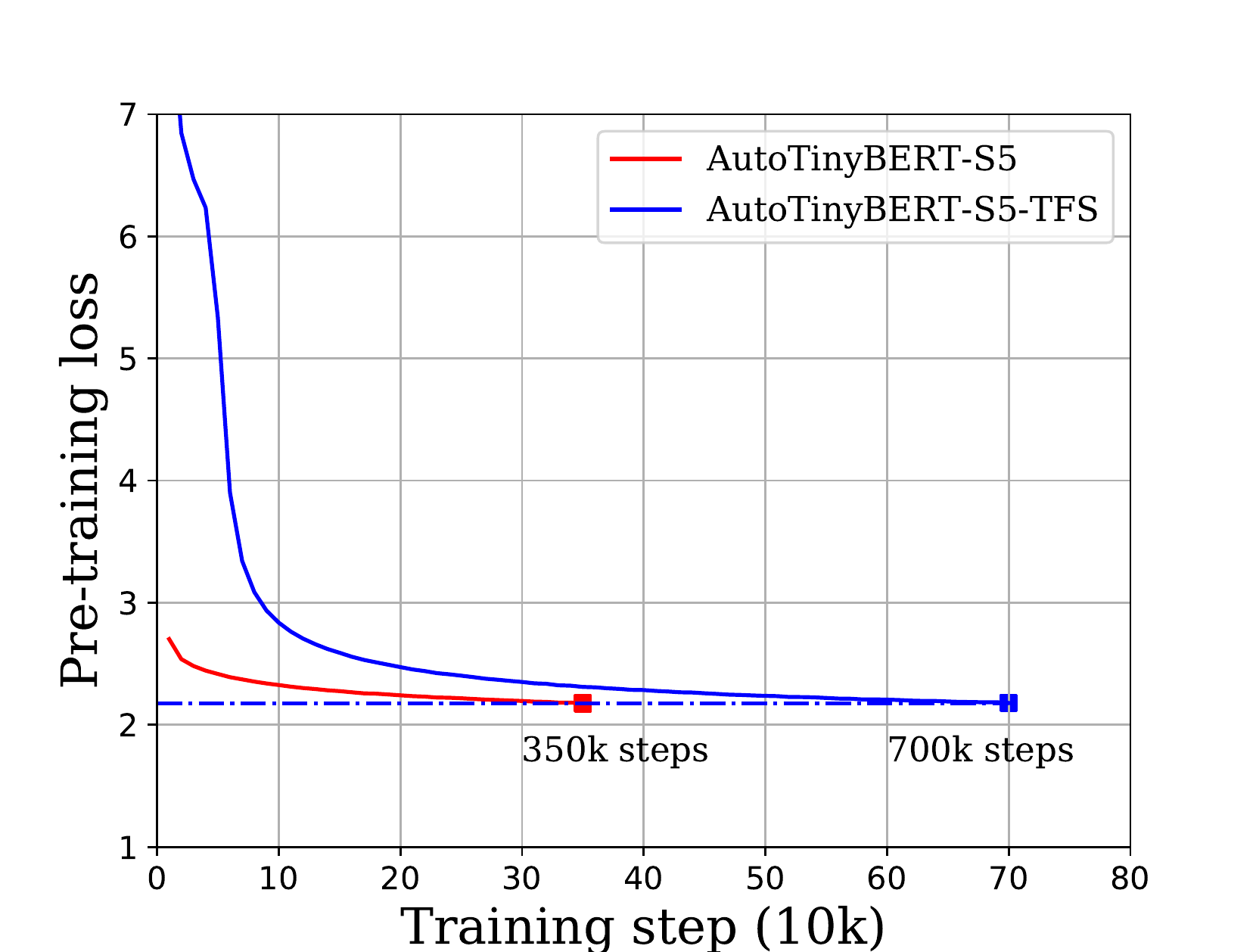}
\caption{Learning curves of AutoTinyBERT and the stand-alone trained model. TFS means the model trained from scratch. AutoTinyBERT can save 50\% training time compared with the model trained from scratch.}
\centering
\label{fig:learning_curve}
\end{minipage}
\end{figure*}

\vspace{-0.1cm}
\subsection{Fast Development of Efficient PLM}
\label{sec:fast_rule}
In this section, we explore an effective setting rule of hyper-parameters based on the obtained architectures and also discuss the computation cost of the development of efficient PLM. The conventional and new architectures are displayed in Table~\ref{tab:arches}. We observe that AutoTinyBERT follows an obvious rule (except the S3 model) in the speedup constraints that are from 4$\times$ to 30$\times$. The rule is summarized as: \{$1.6d^m \leq d^f \leq 1.9d^m,\ \ 0.7d^m \leq d^{q|k|v} \leq 1.0d^m$\}.

With the above rule, we propose a faster way to build efficient PLM, denoted as AutoTinyBERT-Fast.
Specifically, we first obtain the candidates by the rule, and then select $\alpha^{\rm opt}$ from the candidates. We observe the fact that the candidates of the same layer number seem to have similar shapes and we assume that they have similar performance. Therefore, we only need to test one architecture at each layer number and choose the best one as $\alpha^{\rm opt}$. 

To demonstrate the effectiveness of the proposed method, we evaluate these methods at a new speedup constraint of about 10$\times$ under the pre-training setting. The results are shown in Table~\ref{tab:10x_speedup}. We find AutoTinyBERT is efficient and its development time is twice that of the conventional method (BERT) and the result is improved by about 1.8\%. AutoTinyBERT-Fast achieves a competitive score of 77.6 by only about 50\% of BERT training time. In addition to the proposed search method and fast building rule, one reason for the high efficiency of AutoTinyBERT is that the initialization of SuperPLM helps the model to achieve 2$\times$ the convergence speedup, as illustrated in Figure~\ref{fig:learning_curve}.


\section{Related Work}
{\bf Efficient PLMs with Tiny sizes.} 
There are two widely-used methods for building efficient PLMs: pre-training and model compression. Knowledge distillation (KD)~\cite{Hinton2015DistillingTK,romero2014fitnets} is the most widely studied technique in PLM compression, which uses a teacher-student framework. The typical distillation studies include DistilBERT~\cite{sanh2019distilbert}, BERT-PKD~\cite{sun2019patient}, MiniLM~\cite{wang2020minilm}, MobileBERT~\cite{sun12020mobilebert}, MiniBERT~\cite{tsai2019small} and ETD~\cite{chen2021extract}. In addition to KD, the techniques of pruning~\cite{han2015deep_compression,hou2020dynabert}, quantization~\cite{shen2020q,zhang2020ternarybert,wang2020structured} and parameter sharing~\cite{lan2019albert} introduced for PLM compression. Our method is orthogonal to the building method of efficient PLM and is trained under the settings of pre-training and task-agnostic BERT distillation, which can be used by direct fine-tuning.

\vspace{0.1cm}
\noindent{\bf NAS for NLP.} NAS is extensively studied in computer vision~\cite{tan2019efficientnet,tan2020efficientdet}, but relatively little studied in the natural language processing. Evolved Transformer~\cite{so2019evolved} and HAT~\cite{wang2020hat} search architecture for Transformer-based neural machine translation. For BERT distillation, AdaBERT~\cite{chen2020adabert} focuses on searching the architecture in the fine-tuning stage and relies on data augmentation to improve its performance. schuBERT~\cite{khetan2020schubert} obtains the optimal structures of PLM by a pruning method. A work similar to ours is NAS-BERT~\cite{xu2021taskagnostic}. It proposes some techniques to tackle the challenging exponential search space of non-identical layer structure and heterogeneous modules. Our method adopts a linear search space and introduces several practical techniques for SuperPLM training. Moreover, our method is efficient in terms of computation cost and the obtained PLMs are easy to use. 

\section{Conclusion}
\label{sec:conclusion}
We propose an effective and efficient method AutoTinyBERT to search for the optimal architecture hyper-parameters of efficient PLMs. We evaluate the proposed method in the scenarios of both the pre-training and task-agnostic BERT distillation. The extensive experiments show that AutoTinyBERT can consistently outperform the baselines under different latency constraints. Furthermore, we develop a fast development rule for efficient PLMs which can build an AutoTinyBERT model even with less training time of a conventional one. 
\section*{Acknowledgments}

We thank all the anonymous reviewers for their valuable comments. We thank MindSpore\footnote{MindSpore. \url{https://www.mindspore.cn/}} for the partial support of this work, which is a new deep learning computing framework.

\bibliographystyle{acl_natbib}
\bibliography{acl2021}

\clearpage

\appendix

\section{Code Modifications for AutoTinyBERT.}
\label{appendix:code}
We modify the original code\footnote{\url{https://github.com/huggingface/transformers}} to load AutoTinyBERT model and present the details of code modifications in the Figure~\ref{appedix:code_changes}. We assume that $d^{q/k}=d^v$, and more complicated setting is that $d^v$ can be different with $d^{q/k}$, we can do corresponding changes based on the given modifications.

\section{Search Space of Architecture Hyper-parameters.}
\label{appendix:space}
We has trained two SuperPLMs with a architecture of \{$l^t$=$8$, $d^{m/q/k/v}$=$768$, $d^f$=$3072$\} to cover the two scenarios of building efficient PLMs (pre-training and task-agnostic BERT distillation). 
The sampling space in the SuperPLM training is the same as the search space in the search process, as shown in the Table~\ref{tab:search_space}. It can be inferred from the table that the search spaces of the pre-training setting and the knowledge distillation setting are about 46M and 10M, respectively.

\begin{table}[ht]
\centering
	\scalebox{1}{
\begin{tabular}{c|c}
\toprule
Variables $\alpha$       & Search Space $A$                          \\ \midrule
\multicolumn{2}{l}{\it SuperPLM with Pre-training Loss}             \\ \midrule
$l^t$              & {[}1,2,3,4,5,6,7,8{]}                     \\
$d^{m/o}$        & {[}128,132,...,4$k$,...,764,768{]}        \\
$d^{f}$          & {[}128,132,...,4$k$,...,3068,3072{]}      \\
$h$              & {[}1,2,...,$k$,...,11,12{]}               \\
$d^{q/k/v}$      & 64$h$                                     \\ \midrule    
\multicolumn{2}{l}{\it SuperPLM with Knowledge Distillation Loss}   \\  \midrule
$l^t$              & {[}1,2,3,4,5,6,7,8{]}                     \\
$d^{m/o}$        & {[}128,132,...,4$k$,...,764,768{]}        \\
$d^{f}$          & {[}128,132,...,4$k$,...,3068,3072{]}      \\
$h$              & {[}12{]}                        \\ 
$d^{q/k/v}$      & {[}180,192,...,12$k$,...,756,768{]}                      \\ \midrule
\end{tabular}
}
\caption{The search space for architecture hyper-parameters. We assume that $d^{q|k}=d^v$ in the experiment for the training and search efficiency.}
\label{tab:search_space}
\end{table}

\begin{table}[ht]
\centering
	\scalebox{1}{
\begin{tabular}{c|c}
\toprule
Variables $\alpha$       & Search Space $A$                          \\ \midrule
\multicolumn{2}{l}{\it SuperPLM with Knowledge Distillation Loss}   \\  \midrule
$l^t$              & {[}1,2,3,4,5,6,7,8{]}                     \\
$d^{m/o}$        & {[}128,132,...,4$k$,...,764,768{]}        \\
$d^{f}$          & {[}128,132,...,4$k$,...,3068,3072{]}      \\
$h$              & {[}12{]}                        \\ 
$d^{q/k/v}$      & {[}180,192,...,12$k$,...,756,768{]}                      \\ \midrule
\end{tabular}
}
\caption{The search space for architecture hyper-parameters. We assume that $d^{q|k}=d^v$ in the experiment for the training and search efficiency.}
\label{tab:search_space}
\end{table}

\begin{figure*}
\begin{lstlisting}
class BertSelfAttention(nn.Module):
    def __init__(self, config):
        ### Before modifications:
        self.attention_head_size = int(config.hidden_size / config.num_attention_heads)
        ### After modifications: 
        try:
            qkv_size = config.qkv_size
        except:
            qkv_size = config.hidden_size

        self.attention_head_size = int(qkv_size / config.num_attention_heads)

class BertSelfOutput(nn.Module):
    def __init__(self, config):
        ### Before modifications:
        self.dense = nn.Linear(config.hidden_size, config.hidden_size)
        
        ### After modifications:
        try:
            qkv_size = config.qkv_size
        except:
            qkv_size = config.hidden_size

        self.dense = nn.Linear(qkv_size, config.hidden_size)
        
\end{lstlisting}

\caption{Code Modifications to load AutoTinyBERT.}
\label{appedix:code_changes}
\end{figure*}

\begin{algorithm*}
	\caption{The Evolutionary Algorithm}
	\begin{algorithmic}[1]
		\State \textbf{Input}: the number of generations $T=4$, the number of archtectures $\alpha$s in each generation $S=25$, the
		mutation ${\rm Mut(*)}$ probability $p_{m}=1/2$, the exploration probability $p_{e}=1/2$.
		\State Sample first generation $\mathbb{G}_{1}$ from $\mathcal{A}$, and Evoluator produces its performance $\mathbb{V}_1$.
		\For{$t=2, 3 \cdots, T$}
		\State $\mathbb{G}_{t} \gets  \{\} $
		\While{$|\mathbb{G}_{t}|<S$}
		\State Sample one architecture: ${\alpha}$ with a Russian roulette process on $\mathbb{G}_{t-1}$ and $\mathbb{V}_{t-1}$.
		\State With probability $p_{m}$, do ${\rm Mut(*)}$ for ${\alpha}$.
		\State With probability $p_e$, sample a new architecture from $\mathcal{A}$.
		\State Append the newly generated architectures into $\mathbb{G}_{t}$.
		\EndWhile
		\State Evaluator obtains $\mathbb{V}_{t}$  for $\mathbb{G}_{t}$.
		\EndFor
		\State \textbf{Output}: Output the $\alpha^{opt}$ with best performance in the above process. 
	\end{algorithmic}
	\label{alg:ga}
\end{algorithm*}

\section{Evolutionary Algorithm.}
\label{appendix:evo_alg}
We give a detailed description of evolutionary algorithm in Algorithm~\ref{alg:ga}.

\section{Hyper-parameters for Fine-Tuning.}
\label{appendix:hyper-parameter}
Fine-tuning hyper-parameters of GLUE benchmark and SQuAD are displayed in Table~\ref{tab:hyper-params}. AutoTinyBERT and baselines follow the same settings.

\begin{table}[h]
	\centering
	\scalebox{1}{
\begin{tabular}{lccc}
\toprule
\multirow{2}{*}{Tasks} & \multicolumn{1}{c}{Batch} & \multicolumn{1}{c}{Learning} & \multicolumn{1}{c}{\multirow{2}{*}{Epochs}} \\
                       & \multicolumn{1}{c}{size}  & \multicolumn{1}{c}{rate}     & \multicolumn{1}{c}{}                        \\ \midrule
SQuAD                  & 16                        & 3e-5                         & 4                                           \\
SST-2                  & 32                        & 2e-5                         & 4                                           \\
MNLI                   & 32                        & 3e-5                         & 4                                           \\
MRPC                   & 32                        & 2e-5                         & 10                                          \\
CoLA                   & 32                        & 1e-5                         & 10                                          \\
QNLI                   & 32                        & 2e-5                         & 10                                          \\
QQP                    & 32                        & 2e-5                         & 5                                           \\
STS-B                  & 32                        & 3e-5                         & 10                                          \\
RTE                    & 32                        & 2e-5                         & 10                                          \\ \midrule
\end{tabular}
}
\caption{Hyper-parameters used for fine-tuning on GLUE benchmark and SQuAD.}
\label{tab:hyper-params}
\end{table}

\section{Pairwise Accuracy.} 
\label{appendix:pair_acc}
We denote a set of architectures $\{\alpha_1, \alpha_2, ..., \alpha_n\}$ as $\mathcal{A}_{eva}$ and evaluate SuperPLM on this set. The pairwise accuracy is formulated as bellow:

\begin{equation}
\begin{aligned}
\frac{\sum_{\alpha_1\in \mathcal{A}_{eva}, \alpha_2\in \mathcal{A}_{eva}}\mathbbm{1}_{f(\alpha_1)\geq f(\alpha_2)}\mathbbm{1}_{s(\alpha_1)\geq s(\alpha_2)}}{\sum_{\alpha_1\in \mathcal{A}_{eva}, \alpha_2\in \mathcal{A}_{eva}} 1}, 
\end{aligned}
\end{equation}
where $\mathbbm{1}$ is the 0-1 indicator function, $f(*)$ and $s(*)$ refer to the performance of one-shot model and stand-alone trained model respectively.

\section{More details for Fast Development of efficient PLM.}
\label{appendix:details_fast}
We present the detailed results and architecture hyper-parameters for fast development of efficient PLM in Table~\ref{tab:fast_results}.

\begin{table*}[]
	\centering
	\scalebox{0.70}{
		\begin{tabular}{l|c|ccccccccc|c}
			\toprule
			Model  & Speedup                           & SQuAD & SST-2 & MNLI & MRPC & CoLA & QNLI & QQP & STS-B & RTE & Score \\ \midrule
            BERT-S5$_{4-384-1536-6-384}$  & 9.3$\times$  & 78.5  & 86.1 & 76.8  & 83.1 & 35.5 & 84.6 & 87.5 & 86.9 & 65.7 & 76.0 \\
		 AutoTinyBERT-S5$_{5-450-636-6-384}$ & 10.8$\times$  & 79.7   & 89.1 & 78.3 & 84.6  & 39.0 & 85.9 & 88.2 & 87.4 & 68.7 & 77.8 \\
     AutoTinyBERT-Fast-S5$_{5-432-720-6-384}$ & 10.3$\times$ & 80.0  & 88.2  & 77.9  & 84.6 & 37.7 & 86.1 & 88.0 & 87.3 & 68.7 & 77.6  \\ \midrule
\end{tabular}}
\caption{Detailed results for fast development of efficient PLM.}
\label{tab:fast_results}
\end{table*}


\end{document}